\documentclass[sigconf]{acmart}
\AtBeginDocument{%
  }

\copyrightyear{2026}
\acmYear{2026}
\setcopyright{cc}
\setcctype{by}
\acmConference[MM '26]{Proceedings of the 34th ACM International Conference on Multimedia}{November 10--14, 2026}{Rio de Janeiro, Brazil}
\acmBooktitle{Proceedings of the 34th ACM International Conference on Multimedia (MM '26), November 10--14, 2026, Rio de Janeiro, Brazil}
\acmDOI{10.1145/3767308.3835881}
\acmISBN{979-8-4007-2213-4/2026/11}

\usepackage{multirow}

\begin{document}

\title{Graph-Supervised Hierarchical Clinical Alignment for Radiology Report Generation with Large Language Models}

\author{Yingshu Li}
\authornote{Both authors contributed equally to this research.}
\affiliation{%
  \institution{University of Sydney}
  \city{Sydney}
  \country{Australia}}
\email{yili7216@uni.sydney.edu.au}

\author{Yunyi Liu}
\authornotemark[1]
\affiliation{%
  \institution{University of Sydney}
  \city{Sydney}
  \country{Australia}}
\email{yunyi.liu1@sydney.edu.au}

\author{Zhanyu Wang}
\affiliation{%
  \institution{University of Sydney}
  \city{Sydney}
  \country{Australia}}
\email{zhanyu.wang@sydney.edu.au}

\author{Zailong Chen}
\affiliation{%
  \institution{University of Wollongong}
  \city{Wollongong}
  \country{Australia}}
\email{zc881@uowmail.edu.au}

\author{Lingqiao Liu}
\affiliation{%
  \institution{University of Adelaide}
  \city{Adelaide}
  \country{Australia}}
\email{lingqiao.liu@adelaide.edu.au}

\author{Lei Wang}
\affiliation{%
  \institution{University of Wollongong}
  \city{Wollongong}
  \country{Australia}}
\email{leiw@uow.edu.au}

\author{Luping Zhou}
\correspondingauthor
\affiliation{%
  \institution{University of Sydney}
  \city{Sydney}
  \country{Australia}}
\email{luping.zhou@sydney.edu.au}

\renewcommand{\shortauthors}{Yingshu Li et al.}

\begin{abstract}
Radiology report generation (RRG) has recently benefited from large language models, which substantially improve report fluency. However, clinically faithful generation remains challenging because current supervision is still imposed mostly at the report level. This creates a granularity mismatch: radiology reports are composed of disease-grounded findings, while existing methods are trained mainly with whole-report objectives. To address this problem, we propose Graph-Supervised Hierarchical Clinical Alignment, which reformulates image-report supervision as a hierarchical clinical alignment problem. Our method structures this alignment as a disease-conditioned process, where supervision is decomposed into two levels: Disease-Centric Alignment for fine-grained disease-specific correspondence, and Global Clinical Semantic Alignment for report-level semantic coherence. A clinical knowledge graph is used as a training-time-only structural prior that defines disease-specific supervision units and their clinical relationships, introducing no additional overhead at inference. Because standard contrastive alignment could produce false negatives when studies share overlapping pathologies, we combine instance-conditioned discriminative matching with disease-conditioned soft regularization, enabling fine-grained yet clinically consistent cross-modal representations. Experiments on MIMIC-CXR, IU-Xray, and COV-CTR show that our method consistently improves performance on both conventional and clinical metrics. Notably, our 3B model surpasses several prior systems with larger 7B/13B backbones, suggesting that improving supervision structure, rather than increasing model size, can be more effective for RRG.
\end{abstract}




\maketitle

\section{Introduction}
\label{sec:introduction}

\begin{figure}[t]
\centering
\centerline{\includegraphics[width=\linewidth]{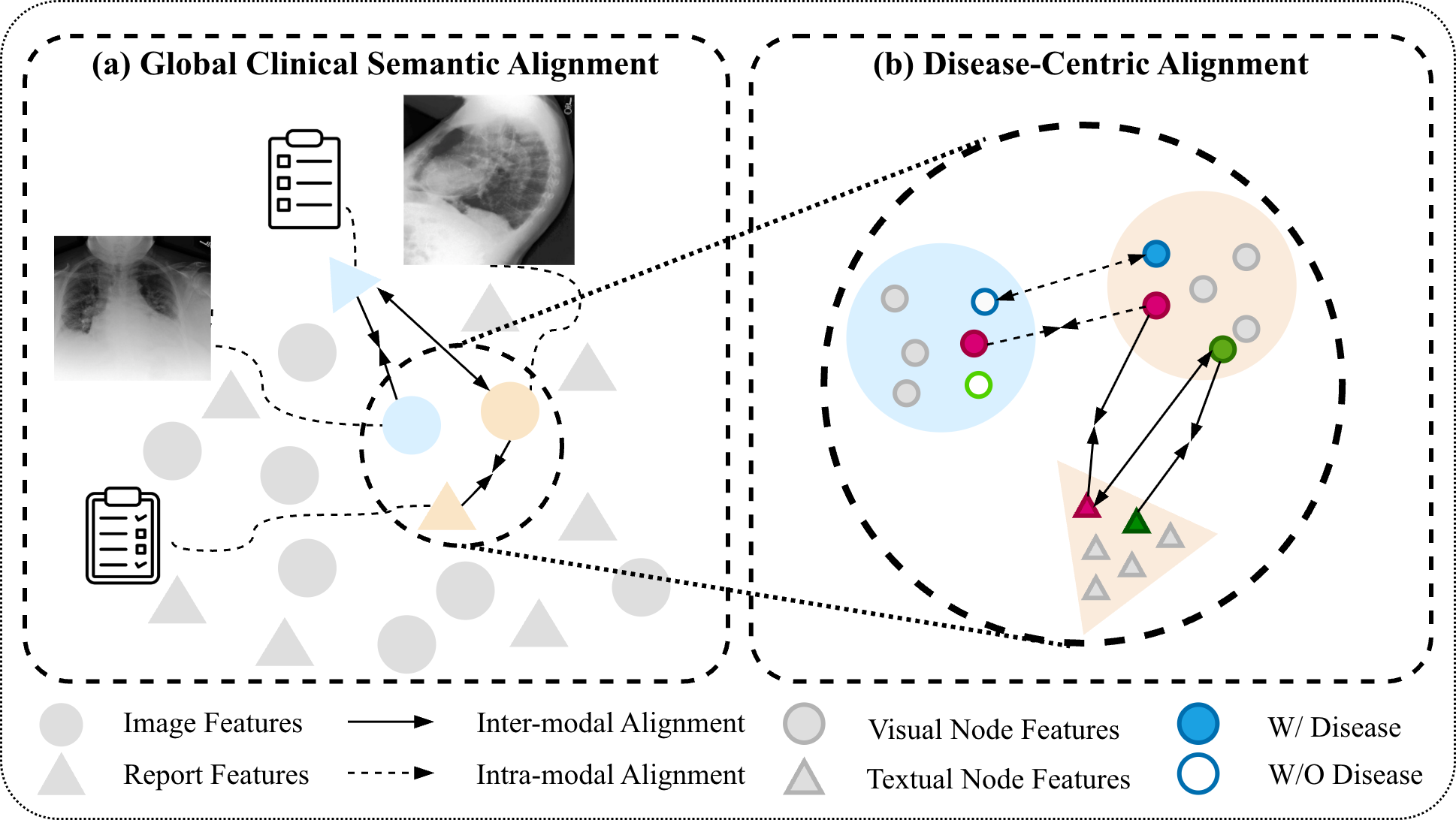}}
\caption{
Our framework decomposes image-report supervision into two complementary levels: (a) \textbf{Global Clinical Semantic Alignment}, which preserves holistic semantic consistency between the image and report; and (b) \textbf{Disease-Centric Alignment}, which enforces fine-grained alignment of clinically relevant visual and textual features.
}
\vspace{-6mm}
\label{fig:story}
\end{figure}
Radiology report generation (RRG) has advanced rapidly with large language models (LLMs) and multimodal LLMs (MLLMs), which substantially improve report fluency~\cite{dubey2024llama, yang2024qwen2, zhu2023minigpt, liu2024visual, ye2024mplug, nicolson2023improving, wang2023r2gengpt, liu2024bootstrapping, li2024kargen, pellegrini2023radialog}. As a result, the current bottleneck of RRG lies not in language generation capacity, but in the granularity of supervision. A radiology report is not a single semantic target: it is a structured composition of disease-specific findings, each grounded in distinct visual evidence. Current RRG models, however, are still trained mainly with holistic report-level objectives, creating a \emph{supervision granularity mismatch}: clinically meaningful semantics are local and disease-conditioned, while training signals remain global and entangled. As a result, token-level supervision conflates multiple findings into a single training signal, weakening disease-specific alignment and limiting precise clinical correspondences.

Existing efforts fall into two categories, neither of which resolves this mismatch. The first category strengthens representations or generation architectures, including transformer-based encoder-decoders~\cite{vaswani2017attention, chen2020generating, chen2022cross, wang2022automated, huang2023kiut, bu2024instance}, contrastive image-text matching~\cite{wang2022automated, li2023dynamic}, and CLIP-based cross-modal models~\cite{radford2021learning, wang2022medclip}. They improve report quality, but supervision remains at the sample level. The second category injects clinical priors such as disease labels~\cite{wang2022medical}, disease relation graphs~\cite{liu2021exploring, huang2023kiut}, or normal-abnormal contrastive cues~\cite{li2025contrastive}, but uses them to enrich features rather than to restructure the granularity at which supervision is imposed. In both cases, holistic training signals remain structurally misaligned with finding-level clinical semantics.

We argue that clinically faithful report generation demands supervision whose granularity matches the clinical structure of the output. A radiology report conveys two levels of semantics simultaneously: individual disease findings that correspond to localized visual evidence, and holistic report coherence that integrates these findings into a consistent clinical narrative. Neither level alone is sufficient: finding-level supervision without report-level coherence produces fragmented descriptions, while report-level supervision without disease-level disentanglement conflates co-occurring findings. This hierarchy is not a design choice imposed by the model; it is induced by the clinical structure of radiology reports themselves, requiring supervision that decomposes image-report alignment into fine-grained disease-conditioned correspondence and global report-level semantic alignment.

To address this mismatch, we propose \textbf{Graph-Supervised Hierarchical Clinical Alignment} (Fig.~\ref{fig:story}), a training framework that restructures image-report supervision in RRG as a two-level clinical alignment problem. The core idea is to use a clinical knowledge graph to define disease-specific supervision units, so that holistic image-report learning is decomposed into clinically meaningful subproblems, each corresponding to a specific disease finding and its visual evidence. At the fine-grained level, these graph-defined units enable \textbf{Disease-Centric Alignment}, which establishes cross-modal correspondence at the granularity of individual findings rather than whole reports. At the global level, \textbf{Global Clinical Semantic Alignment} preserves report-level coherence by aligning image and report representations in a shared semantic space. Because different studies frequently share the same findings, purely instance-wise discriminative matching is complemented by disease-conditioned soft regularization at both levels, preventing clinically similar studies from being over-separated.

The core contribution of this work is not the knowledge graph itself, but the principle that structured clinical relations can redefine how supervision is organized in RRG. In our framework, the knowledge graph serves as a \textbf{training-time structural prior}: it defines the granularity of alignment and the relationships between supervision units, but will be entirely removed after training. Unlike prior graph-based RRG methods that carry knowledge-induced components through the inference pipeline~\cite{li2024kargen, huang2023kiut, liu2021exploring, zhang2020radiology}, our model retains the same inference architecture as the base generator with no additional graph computation. Despite using a smaller 3B backbone, our method achieves comparable or superior performance to larger 7B and 13B systems, suggesting that the bottleneck in current RRG is not model capacity but supervision structure.

\underline{Our contributions are as follows:}

\noindent(1) We identify \emph{supervision granularity mismatch} as a key bottleneck of current RRG systems and reformulate image-report supervision as a \textbf{hierarchical clinical alignment} problem that captures both disease-level correspondence and report-level coherence.

\noindent(2) We propose \textbf{Graph-Supervised Hierarchical Clinical Alignment}, which uses a knowledge graph as a \textbf{training-time structural prior} to factorize supervision into \textbf{Disease-Centric Alignment} and \textbf{Global Clinical Semantic Alignment}, introducing no inference-time graph computation.

\noindent(3) Experiments on three benchmarks show that our 3B model achieves state-of-the-art or competitive results on both conventional and clinical metrics, outperforming several 7B/13B baselines.

\section{Related Work}
\label{sec:relate}
\subsection{Radiology Report Generation (RRG)}
RRG aims to generate clinically accurate and semantically coherent reports from radiological images~\cite{wang2023metransformer, li2024kargen, huang2023kiut, bu2024instance, li2023comprehensive, li2026seeing, li2025s}. Non-LLM-based approaches mainly improve visual feature extraction or incorporate auxiliary clinical knowledge. For example, METransformer~\cite{wang2023metransformer} enhances cross-attention with expert tokens, EKAGen~\cite{bu2024instance} uses instance-level expert knowledge, and KiUT~\cite{huang2023kiut} injects clinical knowledge through a U-Transformer. While these methods incorporate disease-relevant priors, they mainly strengthen representations or generation architectures rather than redesigning image-report supervision. Large language models (LLMs)~\cite{dubey2024llama, yang2024qwen2} have recently been introduced into RRG to improve generation quality. R2GenGPT~\cite{wang2023r2gengpt} employs LLaMA2-7B with a linear projection layer to connect visual and textual features, while MiniGPT-4~\cite{zhu2023minigpt} has been adapted for RRG~\cite{liu2024bootstrapping} with in-domain instance induction and coarse-to-fine decoding. Despite their improved fluency, these models still rely largely on holistic supervision and lack mechanisms to explicitly enforce disease-specific image-report alignment.

\subsection{Feature Alignment in RRG}
Feature alignment in RRG typically relies on contrastive or matching-based objectives to reduce the image--report modality gap. Inspired by vision-language pretraining~\cite{li2022blip}, DCL~\cite{li2023dynamic} contrasts global representations,~\cite{wang2022automated} uses a temporal-weighted matching loss to counter image-text imbalance, and~\cite{huang2024knowledge} aligns globally after feature refinement. These methods improve cross-modal consistency but operate at the holistic or sample level, too coarse for disease-specific correspondences. Finer granularity has been explored---\cite{liu2024multi} introduces sentence-level alignment---yet without disease-structured guidance. What remains missing is an alignment framework that is simultaneously fine-grained and organized around disease-conditioned clinical semantics.

\subsection{Knowledge Graph in RRG}
Structured disease knowledge has also been incorporated into RRG and vision-language learning. IGCL~\cite{khanna2023learning} exploits graph structure for global image-graph alignment during pre-training, but is not designed for report generation. KARGEN~\cite{li2024kargen}, KiUT~\cite{huang2023kiut}, and EKAGen~\cite{bu2024instance} inject disease knowledge through knowledge-guided decoding, prompt generation, or expert retrieval, each adding knowledge-related components to the inference pipeline. Our method instead treats the knowledge graph purely as a training-time structural prior defining disease-specific supervision units and their clinical relationships: all graph components are removed at inference, preserving disease structure in the learned representations at zero additional cost.

\begin{figure*}[t]
\centering
\centerline{\includegraphics[width=0.9\linewidth]{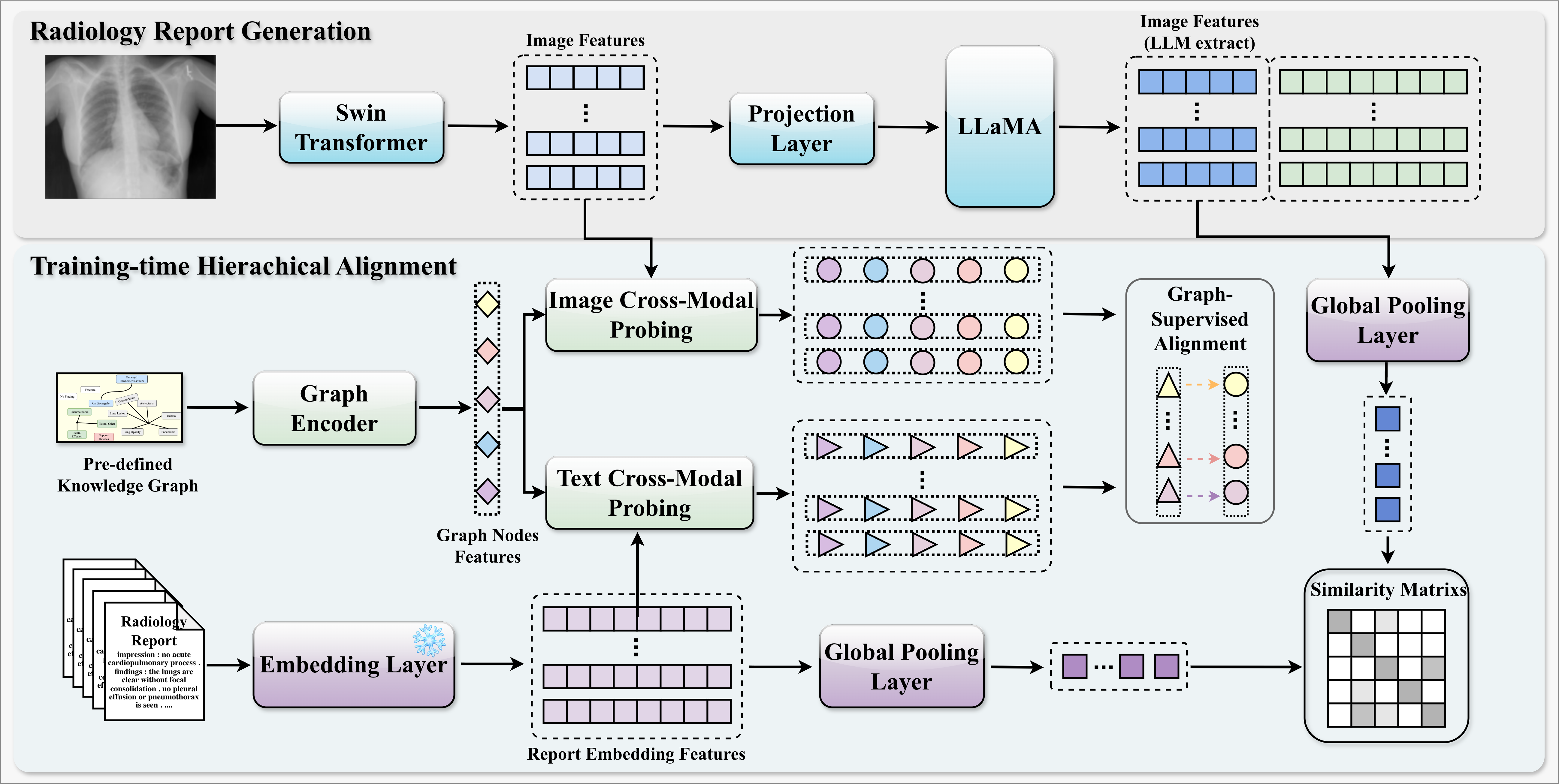}}
\caption{
An overview of our framework. The upper branch is the base \textbf{Radiology Report Generation (RRG)} pipeline. The lower branch is the proposed \textbf{training-time hierarchical clinical alignment}: a knowledge graph defines disease-specific supervision units for \textbf{Disease-Centric Alignment}, while \textbf{Global Clinical Semantic Alignment} preserves report-level coherence. At inference, all graph-related modules are removed; only the RRG backbone is retained.
}
\label{fig:framework}
\end{figure*}


\section{Method}
\label{sec:method}

Radiology report generation (RRG) is typically trained with token-level maximum likelihood, yet clinical correctness is defined over individual findings rather than whole reports. To correct this supervision granularity mismatch, we propose \textbf{Graph-Supervised Hierarchical Clinical Alignment}, which uses a clinical knowledge graph to define disease-specific supervision units during training, decomposing image-report alignment into two levels: \textbf{Disease-Centric Alignment} for fine-grained disease-conditioned correspondence, and \textbf{Global Clinical Semantic Alignment} for report-level coherence. Figure~\ref{fig:framework} gives an overview of the proposed framework.
\vspace{-5mm}
\subsection{Radiology Report Generation}

Our base RRG model follows a standard vision-language pipeline. Given a batch of image-report pairs $\{(\mathbf{I}_i,\mathbf{T}_i)\}_{i=1}^B$, each image is encoded into visual tokens, projected into the LLM embedding space, and used to condition autoregressive report generation. Our hierarchical clinical alignment objectives are imposed on top of this backbone during training.

\noindent\textbf{Vision Encoder.}~~We use a pre-trained Swin Transformer~\cite{liu2021swin} as the visual encoder, which extracts visual token features $\mathbf{F}_{vi} \in \mathbb{R}^{L_v \times d_v}$ from the input image $\mathbf{I}_i$. A two-layer MLP with GELU and layer normalization projects $\mathbf{F}_{vi}$ into the LLM embedding space as $\tilde{\mathbf{F}}_{vi}$. For the alignment objectives introduced later, we also extract report-side text embeddings $\mathbf{G}_i = \text{E}(\mathbf{T}_i)$ from the ground-truth report.

\noindent\textbf{Report Generator.}~~We adopt LLaMA3-3B as the language backbone. The projected visual tokens $\tilde{\mathbf{F}}_{vi}$ are concatenated with an instruction prompt $\mathbf{T}_p$ and fed into the LLM for autoregressive report generation. During training, the generator is optimized with the standard token-level cross-entropy objective:
\begin{equation}
\mathcal{L}_{\mathrm{CE}} =
\mathbb{E}_{(\mathbf{x},\mathbf{y})\sim\mathcal{D}}
\left[
\sum_{j=1}^{N_r}
-\log p_\theta(y_j \mid \mathbf{x}, y_{<j})
\right],
\end{equation}
where $\mathbf{x} = (\tilde{\mathbf{F}}_{vi}, \mathbf{T}_p)$ denotes the multimodal conditioning input, and $\mathbf{y} = (t_1^*, \ldots, t_{N_r}^*)$ is the ground-truth report. While this objective ensures fluent generation, it does not explicitly align visual evidence with clinically meaningful semantics. We therefore impose additional hierarchical clinical alignment objectives during training to regularize the learned representation space.

\subsection{Graph-Supervised Hierarchical Clinical Alignment}

Holistic report supervision entangles multiple co-occurring findings into a single sample-level representation, making fine-grained disease-specific correspondence difficult to learn. To address this limitation, we instantiate the fine-grained level of our framework through \textbf{graph-supervised disease-centric alignment}, which decomposes image-report alignment into disease-conditioned subproblems. Each graph node acts as a clinically meaningful latent query that probes visual and textual representations for evidence associated with a specific disease finding, enabling node-level correspondence beyond holistic report matching. The clinical knowledge graph is used only as a \emph{training-time structural prior}: it specifies how supervision should be factorized into disease-specific units and their clinical relationships, without introducing graph-related computation at inference time. This formulation introduces two node-level alignment problems. The first is cross-modal discrimination: for each disease node, the visual and textual features of the same study should be closer than those of different studies. \textbf{Instance-Conditioned Disease Matching (ICDM)} addresses this by treating the matched image-report pair as the only positive per node. However, strict one-to-one matching produces false negatives when different studies share the same pathology. We thus introduce \textbf{Disease-Conditioned Node Regularization (DCNR)}, which relaxes hard targets using disease labels as soft supervision.

\subsubsection{Clinical Knowledge Graph Construction}
\begin{figure}[h]
\centering
\centerline{\includegraphics[width=0.8\linewidth]{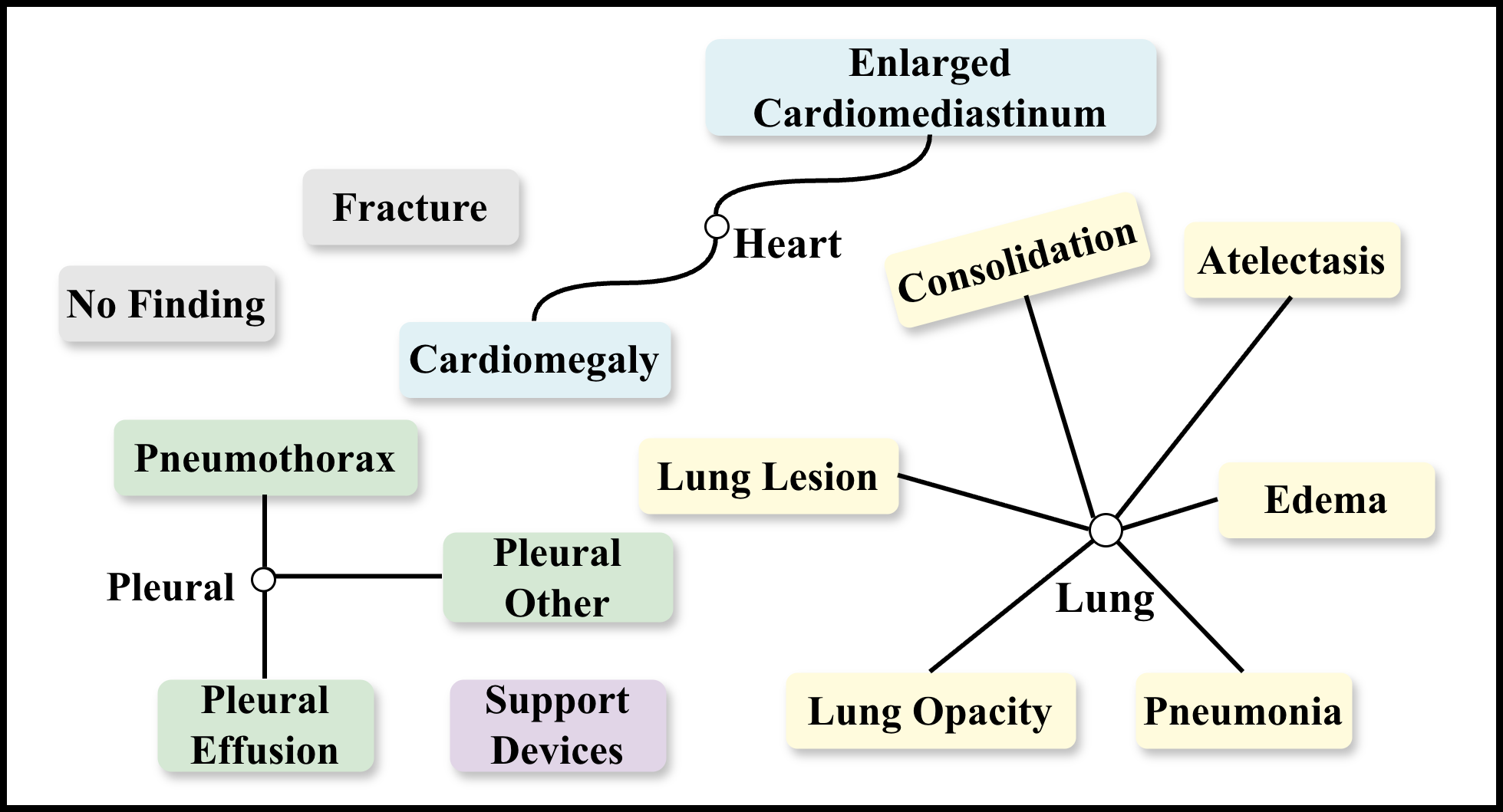}}
\caption{Medical Domain Knowledge Graph.}
\label{fig:graph}
\end{figure}

To support fine-grained clinical alignment, we construct a clinical knowledge graph (Fig.~\ref{fig:graph}) over the 14 CheXpert observation categories~\cite{irvin2019chexpert}. Each category is treated as a graph node, initialized from the LLaMA embedding of its category name. Edges encode anatomical and semantic relatedness: observations associated with the same region (e.g., lung, pleura, or heart) are connected to capture shared imaging representations. In this way, the graph provides a clinical prior that organizes disease concepts before cross-modal alignment is learned.

\subsubsection{Disease-Conditioned Cross-Modal Probing}

Radiology studies often contain multiple co-occurring findings whose visual patterns and textual descriptions are entangled, making global cross-modal interaction too coarse for fine-grained clinical alignment. To address this issue, we introduce \textbf{Disease-Conditioned Cross-Modal Probing}, which uses disease-specific latent queries to retrieve observation-level evidence from both image and text representations. We initialize a set of disease nodes $\mathbf{N}^{0} \in \mathbb{R}^{M \times d_t}$, where $M=14$ denotes the CheXpert observation categories. To encode clinical dependencies among findings, the nodes are refined with a graph encoder under a predefined adjacency matrix $A \in \mathbb{R}^{M \times M}$:
\begin{equation}
\begin{aligned}
\mathbf{N}^{l-1'} &= \mathrm{LayerNorm}\!\left(\mathrm{GSA}(\mathbf{N}^{l-1}, A) + \mathbf{N}^{l-1}\right), \\
\mathbf{N}^{l} &= \mathrm{LayerNorm}\!\left(\mathrm{FFN}(\mathbf{N}^{l-1'}) + \mathbf{N}^{l-1'}\right),
\end{aligned}
\end{equation}
where graph self-attention restricts message passing to clinically related nodes. The refined disease queries $\mathbf{N}^{L}$ then probe modality-specific features via multi-head cross-attention:
\begin{equation}
\mathbf{Z}^{I_i} = \mathrm{MHA}(\mathbf{N}^{L}, \mathbf{F}_{vi}), \qquad
\mathbf{Z}^{T_i} = \mathrm{MHA}(\mathbf{N}^{L}, \mathbf{G}_i).
\end{equation}
This yields disease-level visual and textual representations conditioned on individual clinical entities.


\subsubsection{Instance-Conditioned Disease Matching (ICDM)}

\textbf{ICDM} treats the visual and textual features of the same sample as the only positive pair for each disease node, enforcing disease-specific cross-modal discrimination while preserving instance identity. Given a mini-batch of size $B$, let ${\mathbf z}^{I_i}_m$ and ${\mathbf z}^{T_i}_m$ denote the disease-aware visual and textual features of node $m$ in sample $i$. For each visual node feature ${\mathbf z}^{I_i}_m$, ICDM computes its similarity distribution over the textual node features $\{{\mathbf z}^{T_j}_m\}_{j=1}^B$ of the same disease node across the batch: $p_i({\mathbf z}^{I_i}_m, {\mathbf z}^{T}_m) =
\left[
\frac{\exp(\mathrm{sim}(\tilde{\mathbf z}^{I_i}_m,\tilde{\mathbf z}^{T_j}_m)/\tau)}
{\sum_{k=1}^B \exp(\mathrm{sim}(\tilde{\mathbf z}^{I_i}_m,\tilde{\mathbf z}^{T_k}_m)/\tau)}
\right]_{j=1}^B$, where $\tilde{\mathbf z}^{I_i}_m$ and $\tilde{\mathbf z}^{T_j}_m$ are $\ell_2$-normalized features, and $\tau$ is a temperature parameter.

We supervise this distribution with a one-hot target $\mathbf{y}_i^m$, where $y_{ii}^m = 1$ for the matched image-report pair and $y_{ij}^m = 0$ for $j \neq i$. The image-to-text loss for node $m$ is
\begin{equation}
\mathcal{L}_{\text{ICDM-i2t}}^m =
-\frac{1}{B}\sum_{i=1}^B
\mathbf{y}_i^m \log p_i({\mathbf z}^{I_i}_m,{\mathbf z}^{T}_m),
\end{equation}
and a symmetric text-to-image objective $\mathcal{L}_{\text{ICDM-t2i}}^m$ is defined analogously. The overall loss is
\begin{equation}
\mathcal{L}_{\text{ICDM}} =
\frac{1}{2M}\sum_{m=1}^{M}
\left(
\mathcal{L}_{\text{ICDM-i2t}}^m +
\mathcal{L}_{\text{ICDM-t2i}}^m
\right).
\end{equation}

\subsubsection{Disease-Conditioned Node Regularization (DCNR)}

While ICDM provides strong instance-level discrimination, its one-hot supervision can produce false negatives when different studies share the same finding. \textbf{DCNR} relaxes strict one-to-one matching using disease labels as soft supervision. Inspired by SoftCLIP~\cite{gao2024softclip}, we construct a disease-conditioned soft target for each node from the 14 CheXpert labels~\cite{irvin2019chexpert}. For disease node $m$, let $\mathbf{l}_i^m$ denote whether finding $m$ is present in sample $i$. We define a label-induced affinity vector $\mathbf{S}_i^m = [\mathbf{l}_i^m \wedge \mathbf{l}_j^m]_{j=1}^B$, which assigns nonzero affinity to samples sharing the same finding. To preserve instance identity, we add an instance-preserving term $\delta_i^m$, where $\delta_{ij}^m = 1$ if $i=j$ and $0$ otherwise. The resulting soft target is $\tilde{\mathbf{S}}_i^m = \text{Softmax}(\mathbf{S}_i^m + \delta_i^m)$. We then regularize the predicted cross-modal similarity distribution toward this target using KL divergence:
\begin{equation}
\mathcal{L}_{\text{DCNR-i2t}}^m =
\frac{1}{B}\sum_{i=1}^B
\mathrm{KL}\!\left(
\tilde{\mathbf{S}}_i^m
\parallel
p_i({\mathbf z}^{I_i}_m,{\mathbf z}^{T}_m)
\right).
\end{equation}
A symmetric text-to-image objective $\mathcal{L}_{\text{DCNR-t2i}}^m$ is defined analogously. DCNR complements ICDM by allowing clinically similar samples to contribute soft positive signal, making node-level alignment more robust to shared pathologies.

\subsubsection{Intra-Modal Semantic Consistency (IMSC)}

Without intra-modal structure, disease-aware features may still overlap within each modality, weakening disease-specific separability. \textbf{IMSC} addresses this by regularizing intra-modal feature distributions using the same disease-conditioned soft targets. For disease node $m$, IMSC encourages samples sharing the same finding to remain close within each modality, while suppressing similarity to samples with different findings. The visual intra-modal loss is
\begin{equation}
\mathcal{L}_{\text{IMSC-i2i}}^m =
\frac{1}{B}\sum_{i=1}^{B}
\mathrm{KL}
\left(
\tilde{\mathbf{S}}_i^{m}
\parallel
p_i({\mathbf z}^{I_i}_m, {\mathbf z}^{I}_m)
\right),
\end{equation}
where $p_i({\mathbf z}^{I_i}_m, {\mathbf z}^{I}_m) =
\left[
\frac{
\exp(\mathrm{sim}(\tilde{\mathbf z}^{I_i}_m, \tilde{\mathbf z}^{I_j}_m)/\tau)
}{
\sum_{k=1}^{B}
\exp(\mathrm{sim}(\tilde{\mathbf z}^{I_i}_m, \tilde{\mathbf z}^{I_k}_m)/\tau)
}
\right]_{j=1}^{B}$.

A symmetric textual objective $\mathcal{L}_{\text{IMSC-t2t}}^m$ is defined analogously.

Together, the three node-level objectives are complementary: ICDM provides discriminative cross-modal alignment, DCNR relaxes strict one-hot supervision, and IMSC preserves disease-aware semantic structure within each modality. The resulting graph supervised disease centric alignment loss is
\begin{equation}
\mathcal{L}_{\text{GA}} =
\mathcal{L}_{\text{ICDM}}
+
\frac{1}{4M}
\sum_{m=1}^{M}
\left(
\mathcal{L}_{\text{DCNR}}^m
+
\mathcal{L}_{\text{IMSC}}^m
\right).
\end{equation}

\subsection{Global Clinical Semantic Alignment}

Fine-grained disease-level correspondence alone does not guarantee that the image and report remain coherent as a clinical whole, because a radiology report is not merely a set of independent findings but an integrated clinical narrative. We therefore introduce \textbf{Global Clinical Semantic Alignment (GCSA)}, which aligns image and report representations in a shared report-level semantic space. Starting from the projected visual tokens $\tilde{\mathbf{F}}_{vi}$, we feed the visual prompts together with the instruction prompt $\mathbf{T}_p$ into the LLaMA backbone and extract contextualized visual token representations: $\mathbf{V}_{i} = \mathrm{LLaMA}(\tilde{\mathbf{F}}_{vi}, \mathbf{T}_p)$, where $\mathbf{V}_i \in \mathbb{R}^{L_v \times d_t}$. Because these features have already interacted with the language backbone, they capture richer report-level semantic context than raw visual encoder outputs. We then obtain global image and report representations by attention-based pooling: $\mathbf{v}_i = \mathrm{R}_I(\mathbf{V}_{i}) = \sum_{j=1}^{L_v} w_{ij}^{V}\mathbf{V}_{i,j}$, 
$\mathbf{t}_i = \mathrm{R}_T(\mathbf{G}_i) = \sum_{j=1}^{L_t} w_{ij}^{G}\mathbf{G}_{i,j}$, where $\mathbf{w}_{i}^{V}$ and $\mathbf{w}_{i}^{G}$ are attention weights defined by a two-layer MLP with GELU and Softmax. Finally, we compute the similarity distribution between the global image representation $\mathbf{v}_i$ and all textual representations $\{\mathbf{t}_j\}_{j=1}^{B}$ in the batch: $p_i({\mathbf I}_i,{\mathbf T}) =
\left[
\frac{\exp(\mathrm{sim}(\tilde{\mathbf v}_i,\tilde{\mathbf t}_j)/\tau)}
{\sum_{k=1}^B \exp(\mathrm{sim}(\tilde{\mathbf v}_i,\tilde{\mathbf t}_k)/\tau)}
\right]_{j=1}^B$.

\subsubsection{Instance-Conditioned Semantic Matching (ICSM)}

While disease-centric alignment captures fine-grained correspondence, it does not explicitly preserve instance-level alignment in the global semantic space. We therefore introduce \textbf{Instance-Conditioned Semantic Matching (ICSM)} as the discriminative objective of the global branch, encouraging each image to align with its corresponding report while remaining distinguishable from other samples. For the $i$-th image in a mini-batch, 
based on the global similarity distribution $p_i(\mathbf I_i,\mathbf T)$, the image-to-text loss is
\begin{equation}
\mathcal{L}_{\text{ICSM-i2t}}
=
-\frac{1}{B}
\sum_{i=1}^{B}
\mathbf{y}_i
\log p_i({\mathbf I}_i,{\mathbf T}).
\end{equation}
A symmetric text-to-image objective $\mathcal{L}_{\text{ICSM-t2i}}$ is defined analogously, and the final loss is
\begin{equation}
\mathcal{L}_{\text{ICSM}}
=
\frac{1}{2}
\left(
\mathcal{L}_{\text{ICSM-i2t}}
+
\mathcal{L}_{\text{ICSM-t2i}}
\right).
\end{equation}

\subsubsection{Disease-Conditioned Semantic Regularization (DCSR)}

Analogous to DCNR at the node level, we extend disease-conditioned soft supervision to the global semantic space to reduce false negatives among clinically similar studies. Unlike DCNR, which operates per disease node, \textbf{DCSR} constructs a soft target $\tilde{\mathbf S}_i$ from the full CheXpert label profile of each study, allowing samples with similar disease compositions to remain closer while preserving the matched pair as the dominant target. We regularize the predicted global similarity distribution toward this target using KL divergence:
\begin{equation}
\mathcal{L}_{\text{DCSR-i2t}}
=
\frac{1}{B}
\sum_{i=1}^{B}
\mathrm{KL}
\left(
\tilde{\mathbf S}_i
\parallel
p_i(\mathbf{I}_i,\mathbf{T})
\right).
\end{equation}
A symmetric text-to-image objective $\mathcal{L}_{\text{DCSR-t2i}}$ is defined analogously, and the final DCSR loss averages both directions. DCSR complements ICSM by preserving report-level discrimination while reducing false negatives among clinically similar studies.

\subsection{Overall Training Objective and Inference}

The cross-entropy objective supervises token generation, while the alignment objectives shape the representation space at two complementary levels. The overall training objective is
\begin{equation}
\mathcal{L}
=
\mathcal{L}_{\mathrm{CE}}
+
\lambda_1
\left(
\mathcal{L}_{\text{ICSM}}
+
\mathcal{L}_{\text{DCSR}}
\right)
+
\lambda_2
\mathcal{L}_{\text{GA}},
\end{equation}
where $\lambda_1$ and $\lambda_2$ control the strengths of global clinical semantic alignment and graph-supervised disease-centric alignment, respectively. The framework introduces less than $10\%$ additional trainable parameters over the 3B baseline; All alignment objectives are used only during training. Thus, our method retains the same inference architecture as the base model, with no additional graph computation or inference-time overhead. In this way, we improve clinically structured representation learning through training-time supervision restructuring rather than increased inference-time complexity.

\begin{table*}[h]
\caption{
Comparison on MIMIC-CXR (upper) and IU-Xray (lower).
For IU-Xray, we report variants with and without disease-conditioned objectives (pseudo labels from CheXbert). Bold/underline denote best/second-best. $\dagger$: results from original papers. Multi-Grained$^{\dagger}$~\cite{liu2024multi} is reported without post-processing RL. LLM sizes in parentheses.
}
{\small
\begin{tabular*}{\hsize}{@{}@{\extracolsep{\fill}}l|l|cccccc@{}}
\hline
\textbf{Dataset}   & \textbf{Methods}       & \textbf{BLEU-1}         & \textbf{BLEU-2}         & \textbf{BLEU-3}        & \textbf{BLEU-4}         & \textbf{METEOR}        & \textbf{ROUGE} \\ \hline
\multirow{18}{*}{MIMIC-CXR}  
& R2GenCMN$^{\dagger}$~\cite{chen2022cross}                           & 0.353 & 0.218 & 0.148 & 0.106 & 0.142 & -     \\
& PPKED$^{\dagger}$~\cite{liu2021exploring}                           & 0.360 & 0.224 & 0.149 & 0.106 & 0.149 & 0.237 \\
& METransformer$^{\dagger}$~\cite{wang2023metransformer}              & 0.386 & 0.250 & 0.169 & 0.124 & 0.152 & 0.291 \\
& DCL$^{\dagger}$~\cite{li2023dynamic}                                & -     & -     & -     & 0.109 & 0.150 & 0.284 \\
& KiUT$^{\dagger}$~\cite{huang2023kiut}                               & 0.393 & 0.243 & 0.159 & 0.113 & 0.160 & 0.285 \\

& EKAGen$^{\dagger}$~\cite{bu2024instance}                            & \underline{0.419} & 0.258 & 0.170 & 0.119 & 0.157 & 0.287 \\
& PromptMRG$^{\dagger}$~\cite{jin2024promptmrg}                       & 0.398 & -     & -     & 0.112 & 0.157 & 0.268 \\
& CvT2DistilGPT2$^\dagger$~\cite{nicolson2023improving}               & 0.393 & 0.248 & 0.171 & 0.127 & -     & 0.155 \\

& KCAP$^{\dagger}$~\cite{huang2024knowledge}                          & 0.378 & 0.240 & 0.165 & 0.121 & 0.149 & \underline{0.301} \\
& Multi-Grained$^{\dagger}$~\cite{liu2024multi}                       & 0.346 & 0.226 & 0.159 & 0.117 & 0.163 & 0.290 \\ \cline{2-8}
& R2GenGPT (7B)$^{\dagger}$~\cite{wang2023r2gengpt}                        & 0.411 & 0.267 & 0.186 & 0.134 & 0.160 & 0.297 \\
& RaDialog-RG (7B)$^\dagger$~\cite{pellegrini2023radialog}            & 0.346 & -     & -     & 0.095 & 0.140 & 0.271 \\
& Bootstrapping (13B)$^{\dagger}$~\cite{liu2024bootstrapping}               & 0.402 & 0.262 & 0.180 & 0.128 & \textbf{0.175} & 0.291 \\
& KARGEN (7B)$^{\dagger}$~\cite{li2024kargen}                              & 0.417 & \underline{0.274} & \underline{0.192} & \underline{0.140} & 0.165 & \textbf{0.305} \\ \cline{2-8}
& Ours (3B)                                                                & \textbf{0.427} & \textbf{0.279} & \textbf{0.194} & \textbf{0.143} & \underline{0.167} & \textbf{0.305} \\ \hline
\multirow{20}{*}{IU-Xray}  
& R2GenCMN$^{\dagger}$~\cite{chen2022cross}                       & 0.475 & 0.309 & 0.222 & 0.170 & 0.191& 0.375  \\
& METransformer$^{\dagger}$~\cite{wang2023metransformer}          & 0.483 & 0.322 & 0.228 & 0.172 & 0.192 & 0.380 \\
& DCL$^{\dagger}$~\cite{li2023dynamic}                            & -     & -     & -     & 0.163 & 0.193 & -     \\
& PromptMRG$^{\dagger}$~\cite{jin2024promptmrg}                   & 0.401 & -     & -     & 0.098 & 0.160 & 0.281 \\
& CvT2DistilGPT2$^\dagger$~\cite{nicolson2023improving}           & 0.473 & 0.304 & 0.224 & 0.175 & 0.200 & 0.376 \\
& Multi-Grained$^{\dagger}$~\cite{liu2024multi}                   & 0.472 & 0.321 & 0.234 & 0.175 & 0.192 & 0.379 \\ 
& KCAP$^{\dagger}$~\cite{huang2024knowledge}                      & \textbf{0.517} & 0.334 & 0.243 & 0.184 & 0.210 & 0.384 \\ \cline{2-8}
& R2GenGPT (7B)$^{\dagger}$~\cite{wang2023r2gengpt}                    & 0.488 & 0.316 & 0.228 & 0.173 & 0.211& 0.377 \\
& Bootstrapping (13B)$^{\dagger}$~\cite{liu2024bootstrapping}           & 0.499 & 0.323 & 0.238 & 0.184 & 0.208 & 0.390\\
& KARGEN (7B)$^{\dagger}$~\cite{li2024kargen}                              & 0.490 & 0.323 & 0.232 & 0.180 & \underline{0.218} & 0.385 \\ \cline{2-8}
& Ours (3B) (w/o DC \& IM)                              & 0.509 & \underline{0.344} & \underline{0.256} & \underline{0.197} & \underline{0.218} & \underline{0.392} \\
& Ours (3B) (using pseudo labels)                              & \underline{0.510} & \textbf{0.346} & \textbf{0.259} & \textbf{0.199} & \textbf{0.222} & \textbf{0.398} \\ \cline{2-8}
& \multicolumn{7}{c}{Results below are not strictly comparable due to different data partitions or image resolution. For reference only.} \\ \cline{2-8}
& {\color[HTML]{9B9B9B} PPKED$^{\dagger}$~\cite{liu2021exploring}}         
& {\color[HTML]{9B9B9B} 0.483} & {\color[HTML]{9B9B9B} 0.315} & {\color[HTML]{9B9B9B} 0.224} & {\color[HTML]{9B9B9B} 0.168} & {\color[HTML]{9B9B9B} 0.376} & {\color[HTML]{9B9B9B} 0.187} \\ 
& {\color[HTML]{9B9B9B}KiUT$^{\dagger}$~\cite{huang2023kiut}}
& {\color[HTML]{9B9B9B} 0.525} & {\color[HTML]{9B9B9B} 0.360} & {\color[HTML]{9B9B9B} 0.251} & {\color[HTML]{9B9B9B} 0.185} & {\color[HTML]{9B9B9B} 0.242} & {\color[HTML]{9B9B9B} 0.409} \\
& {\color[HTML]{9B9B9B} EKAGen~\cite{bu2024instance}}                                    
& {\color[HTML]{9B9B9B} 0.497} & {\color[HTML]{9B9B9B} 0.339} & {\color[HTML]{9B9B9B} 0.250} & {\color[HTML]{9B9B9B} 0.190} & {\color[HTML]{9B9B9B} 0.210} & {\color[HTML]{9B9B9B} 0.399} \\

\hline
\end{tabular*}
}
\label{Table:ComparisonWithSOTA}
\end{table*}

\section{Experiments}
\label{sec:experiments}

\subsection{Datasets}
\noindent\textbf{MIMIC-CXR.}
MIMIC-CXR dataset~\cite{johnson2019mimic}, the largest publicly available collection of chest radiographs and associated free-text reports, includes 377,110 images and 227,835 reports from 64,588 patients. For fair comparison, we utilized the dataset’s division defined by \cite{chen2020generating}, i.e., 270790 images for training and 3858 for testing.

\noindent\textbf{IU-Xray.}
IU-Xray~\cite{demner2016preparing} comprises 3,955 de-identified radiology reports and 7,470 images. For consistent comparisons, we follow the dataset partitioning of~\cite{chen2020generating} with a train/validation/test split of 7:1:2 and conduct all evaluations on the test set.

\noindent\textbf{COV-CTR.}
{COV-CTR~\cite{li2023auxiliary} is a CT-based radiology report dataset, consisting of 728 lung CT scans (349 COVID-19, 379 non-COVID) paired with diagnostic reports. Following prior work~\cite{li2023auxiliary}, the data are split into train/validation/test set with a ratio of 7:1:2.}

\subsection{Experimental Settings}
\noindent\textbf{Evaluation Metrics.}
We report NLG metrics, BLEU~\cite{papineni2002bleu}, METEOR~\cite{banerjee2005meteor}, ROUGE-L~\cite{lin2004rouge}, and, for clinical quality, RadGraph F1~\cite{jain2021radgraph}, BERTScore~\cite{zhang2019bertscore}, RadCliQ~\cite{yu2023evaluating}, and LLM-based GREEN~\cite{ostmeier2024green} and RateScore~\cite{zhao2024ratescore}, which correlate better with radiologist assessment.

\noindent\textbf{Implementation Details.}
We pair a frozen LLaMA3-3B with a Swin Transformer visual encoder; the disease-centric branch adds a 3-layer Graph Encoder and a 3-layer cross-modal probing module. Training uses two NVIDIA RTX 3090 GPUs (24GB each) at a learning rate of $1 \times 10^{-4}$, with beam search of width 3 at inference to balance efficiency and output quality. The alignment modules add 214M parameters (\(<10\%\) of the 3B baseline) and are dropped at inference, giving about 2 seconds per case, as in the baseline.

\begin{table}[h]
\vspace{-2mm}
\centering
\caption{{Evaluation on COV-CTR for CT report generation.}}
\label{tab:covctr}
\resizebox{\linewidth}{!}{
{
\begin{tabular}{lcccccc}
\hline
\textbf{Methods} 
& \textbf{B@1} & \textbf{B@2} & \textbf{B@3} & \textbf{B@4} & \textbf{CIDEr} & \textbf{ROUGE-L} \\
\hline
CoAtt~\cite{jing2018automatic}        & 0.709 & 0.645 & 0.603 & 0.552 & 0.672 & \textbf{0.748} \\
SAT~\cite{vinyals2015show}          & 0.697 & 0.621 & 0.568 & 0.515 & 0.659 & 0.723 \\
Vision-BERT~\cite{devlin2019bert}  & 0.710 & 0.653 & 0.606 & 0.558 & 0.684 & 0.747 \\
Vision-GPT~\cite{radford2018improving}   & 0.708 & 0.645 & 0.600 & 0.549 & 0.680 & 0.746 \\
ASGK~\cite{li2023auxiliary}    & 0.712 & 0.659 & 0.611 & 0.570 & 0.684 & 0.746 \\
R2Gen~\cite{chen2020generating}                  & 0.703 & 0.613 & 0.543 & 0.486 & - & 0.658 \\
R2Gen-CMN~\cite{chen2022cross}              & 0.710 & 0.625 & 0.557 & 0.493 & - & 0.672 \\
R2Gen-Mamba~\cite{sun2025r2gen}            & 0.715 & 0.629 & 0.565 & 0.510 & - & 0.668 \\
\hline
Ours
& \textbf{0.747} & \textbf{0.679} & \textbf{0.623} & \textbf{0.574} & \textbf{1.285} & 0.734 \\
\hline
\end{tabular}
}}
\vspace{-4mm}
\end{table}
\begin{figure*}[t]
\centering
\centerline{\includegraphics[width=0.92\linewidth]{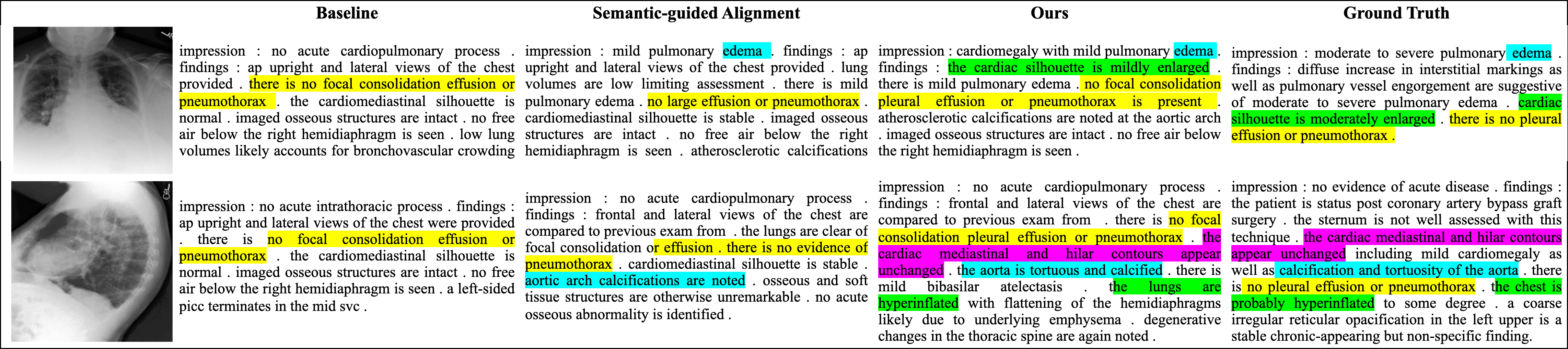}}
\caption{
Examples of generated reports. Different colors correspond to different key disease-related terms.
}
\label{fig:Qulitative}
\end{figure*}

\subsection{Main Results}

Tables~\ref{Table:ComparisonWithSOTA} and~\ref{Table:ComparisonWithSOTA_Rad} report NLG and clinical metrics against encoder-decoder models~\cite{chen2020generating, chen2022cross, liu2021exploring, huang2023kiut, bu2024instance}, contrastive alignment methods~\cite{li2023dynamic, huang2024knowledge, liu2024multi}, and LLM-based approaches~\cite{wang2023r2gengpt, liu2024bootstrapping, pellegrini2023radialog, li2024kargen}. On MIMIC-CXR, our 3B model surpasses R2GenGPT~\cite{wang2023r2gengpt} (7B) and Bootstrapping~\cite{liu2024bootstrapping} (13B), indicating that improving supervision structure can be more effective than scaling the generator. Our method also outperforms Multi-Grained~\cite{liu2024multi} (sentence-level contrastive learning, without RL post-processing) and methods that inject disease knowledge without restructuring supervision, such as EKAGen~\cite{bu2024instance} and KiUT~\cite{huang2023kiut}.

On IU-Xray, disease labels are unavailable, so we first evaluate a reduced variant without disease-conditioned objectives. Even in this setting, the proposed model outperforms most prior methods. Enabling these objectives with CheXbert pseudo labels yields additional gains, suggesting that the framework benefits from clinically structured label signals when available.

On clinical metrics, our method achieves the best RadGraph F1, BERTScore, RadCliQ, GREEN~\cite{ostmeier2024green}, and RateScore~\cite{zhao2024ratescore}, providing evidence that hierarchical clinical alignment improves clinically meaningful report semantics beyond lexical quality. It also outperforms prior methods on COV-CTR (Table~\ref{tab:covctr}), demonstrating cross-modality generalization.

\begin{table}[h]
\caption{Evaluation of Clinic-related Metrics on MIMIC-CXR.}
\centering
\resizebox{\linewidth}{!}{
\begin{tabular}{@{}l|c|c|c|c|c@{}} 
\hline
\textbf{Methods} 
& \textbf{RadGraph F1}($\uparrow$) 
& \textbf{BERTScore}($\uparrow$) 
& \textbf{RadCliQ}($\downarrow$) 
& {\textbf{GREEN($\uparrow$)}} 
& {\textbf{Rate($\uparrow$)}} \\ 
\hline
R2Gen~\cite{chen2020generating}                 
& 0.172 & 0.406 & 1.228 
& {0.276} & {0.526} \\
R2GenCMN~\cite{chen2022cross}
& 0.182 & 0.418 & 1.182
& {0.297} & {\underline{0.538}} \\
CvT2DistilGPT2~\cite{nicolson2023improving}
& 0.196 & 0.374 & 1.220
& {\underline{0.320}} & {0.527} \\
RaDialog-RG$^\dagger$~\cite{pellegrini2023radialog} 
& -     & 0.400 & -     
& {-}     & {-} \\
PromptMRG~\cite{jin2024promptmrg}               
& 0.190 & 0.357 & 1.169 
& {0.287} & {0.528} \\
R2GenGPT~\cite{wang2023r2gengpt}                
& 0.187 & 0.415 & 1.207 
& {0.300} & {0.528} \\
KARGEN~\cite{li2024kargen}                      
& \underline{0.203} & \underline{0.421} & 1.165 
& {0.308} & {0.533} \\
{\color[HTML]{9B9B9B}EKAGen$^{\dagger\dagger}$~\cite{bu2024instance}} 
& {\color[HTML]{9B9B9B}0.199} 
& {\color[HTML]{9B9B9B}0.412} 
& {\color[HTML]{9B9B9B}1.126} 
& {\color[HTML]{9B9B9B}0.256} 
& {\color[HTML]{9B9B9B}0.512} \\
\hline
\textbf{Ours}                                   
& \textbf{0.206} & \textbf{0.422} & \textbf{1.145} 
& {\textbf{0.323}} 
& {\textbf{0.541}} \\
\hline
\midrule[0.2mm] 
\multicolumn{6}{p{13cm}}{\textit{Note: EKAGen uses 300$\times$300 images while the others use 224$\times$224 images.}} \\
\hline
\end{tabular}
}
\label{Table:ComparisonWithSOTA_Rad}
\end{table}

Table~\ref{tab:auroc} reports disease-wise AUROC over the 14 CheXpert categories. Our method shows consistent gains on key findings such as Edema, Atelectasis, Pneumothorax, and Pleural Effusion, consistent with our disease-conditioned alignment design. This reflects improved disease-level semantics beyond holistic fluency.
\begin{table}[h]
\caption{Disease-wise AUROC comparison. 
}
\centering
\small
\setlength{\tabcolsep}{4pt}
\begin{tabular}{lcc|lcc}
\hline
\textbf{Disease} & \textbf{Baseline} & \textbf{Ours} & \textbf{Disease} & \textbf{Baseline} & \textbf{Ours} \\
\hline
Enl. Cardiom.      & 0.488 & \textbf{0.540} & Atelectasis      & 0.629 & \textbf{0.640} \\
Cardiomegaly       & 0.595 & \textbf{0.621} & Pneumothorax     & 0.545 & \textbf{0.551} \\
Lung Opacity       & 0.530 & \textbf{0.567} & Pleural Effusion & 0.766 & \textbf{0.771} \\
Lung Lesion        & 0.503 & \textbf{0.515} & Pleural Other    & 0.502 & \textbf{0.512} \\
Edema              & 0.647 & \textbf{0.676} & Fracture         & 0.505 & \textbf{0.507} \\
Consolidation      & 0.505 & \textbf{0.518} & Support Devices  & 0.706 & \textbf{0.735} \\
Pneumonia          & 0.523 & \textbf{0.572} & No Finding       & 0.619 & \textbf{0.620} \\
\hline
\end{tabular}
\label{tab:auroc}
\end{table}
\subsection{Ablation Study}

\begin{table}[h]
\caption{Component ablation. ICSM: Instance-Conditioned Semantic Matching; DCSR: Disease-Conditioned Semantic Regularization; ICDM: Instance-Conditioned Disease Matching; DCNR: Disease-Conditioned Node Regularization; IMSC: Intra-Modal Semantic Consistency.}
\centering
\resizebox{\linewidth}{!}{
\begin{tabular}{@{}c|c|c|c|c|c|c|c|c|c|c|c@{}}
\hline
\textbf{\#}
& \textbf{Param.}
& \textbf{ICSM}
& \textbf{DCSR}
& \textbf{ICDM}
& \textbf{DCNR}
& \textbf{IMSC}
& \textbf{B@1}
& \textbf{B@4}
& \textbf{METEOR}
& \textbf{ROUGE}
& \textbf{RadCliQ($\downarrow$)} \\
\hline
1 & 7B &  &  &  &  &  & 0.411 & 0.134 & 0.160 & 0.297 & 1.207 \\
2 & 3B &  &  &  &  &  & 0.411 & 0.126 & 0.157 & 0.286 & 1.221 \\

3 & 3B & \checkmark &  &  &  &  & 0.419 & 0.133 & 0.162 & 0.296 & 1.181 \\
4 & 3B &  &  & \checkmark &  &  & 0.416 & 0.131 & 0.161 & 0.293 & 1.195 \\

5 & 3B & \checkmark & \checkmark &  &  &  & 0.424 & 0.135 & 0.165 & 0.296 & 1.177 \\
6 & 3B & \checkmark &  & \checkmark &  &  & 0.421 & 0.136 & 0.164 & 0.294 & 1.179 \\
7 & 3B &  &  & \checkmark & \checkmark &  & 0.420 & 0.134 & 0.164 & 0.294 & 1.175 \\

8 & 3B & \checkmark & \checkmark & \checkmark & \checkmark &  & 0.424 & 0.138 & 0.166 & 0.295 & 1.165 \\
9 & 3B &  &  & \checkmark & \checkmark & \checkmark & 0.425 & 0.139 & 0.166 & 0.299 & 1.155 \\

10 & 3B & \checkmark & \checkmark & \checkmark & \checkmark & \checkmark & \textbf{0.427} & \textbf{0.143} & \textbf{0.167} & \textbf{0.305} & \textbf{1.145} \\
\hline
\end{tabular}
}
\label{tab:ablation}
\end{table}

\subsubsection{Contribution of Each Component}

\begin{figure}[h]
\centering
\centerline{\includegraphics[width=0.9\linewidth]{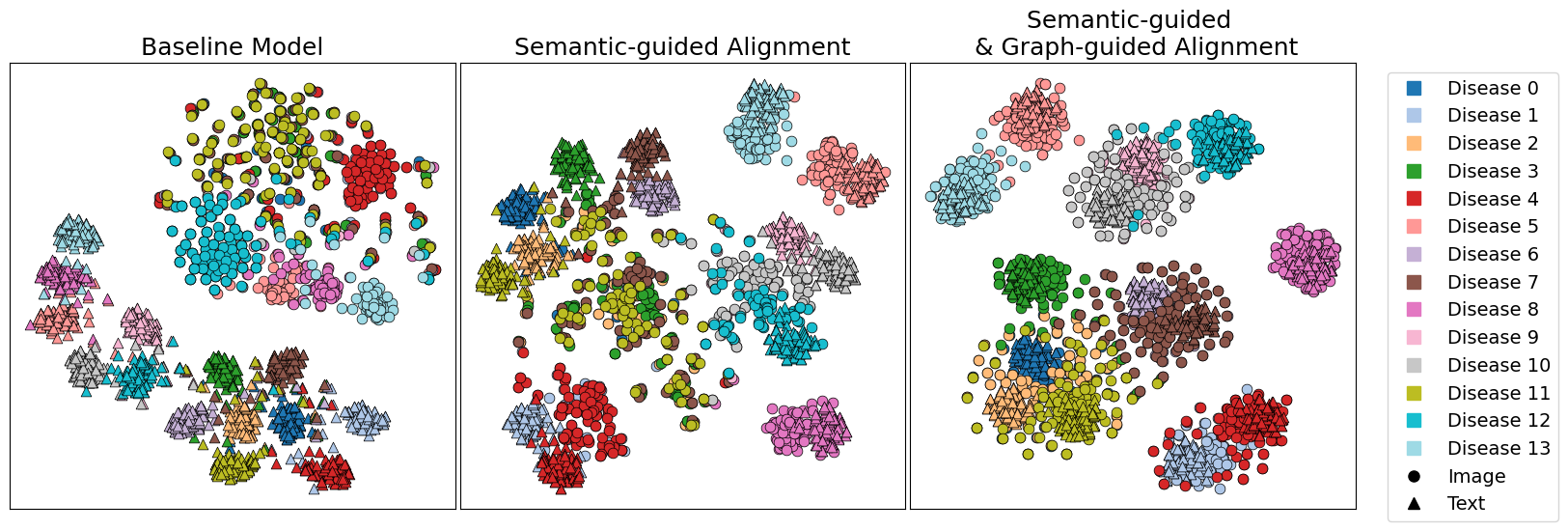}}
\caption{2D projection of disease-level visual (circles) and textual (triangles) features from 100 MIMIC-CXR test samples. Colors denote disease nodes. 
}
\label{fig:vis}
\end{figure}

Table~\ref{tab:ablation} presents the ablation study. The 3B baseline is initially weaker than the 7B baseline; progressively adding our objectives yields consistent improvements across all metrics, and the full model ultimately surpasses the larger backbone. This confirms that clinically structured supervision can be more effective than backbone scaling alone. The ablation further shows why both levels of hierarchical clinical alignment are necessary. Global-only alignment (ICSM+DCSR, row 5) improves report-level correspondence but leaves RadCliQ at 1.177, indicating limited finding-level precision. Node-only alignment (ICDM+DCNR+IMSC, row 9) improves disease-specific separability and reaches a lower RadCliQ of 1.155, but does not fully close the gap on BLEU-1. The full model (row 10) achieves the best performance on all metrics. This pattern reflects the two-level structure of radiology reports themselves: report-level coherence and finding-level precision are both required, and neither level of supervision alone is sufficient to capture both. Figure~\ref{fig:vis} visualizes this complementarity on 100 MIMIC-CXR test samples. Without alignment, the two modalities are scattered. Global Clinical Semantic Alignment reduces the cross-modal gap, and Disease-Centric Alignment further clusters same-disease features while separating different findings.

\subsubsection{Impact of Global Clinical Semantic Alignment}

Introducing \textbf{ICSM} substantially improves the 3B baseline, bringing it close to the 7B model (B@4: 0.126$\to$0.133, a 5.6\% relative gain). This result reveals that a conventional projection layer alone is insufficient to establish strong image-report correspondence in the medical domain; explicit report-level alignment is needed even when the language backbone is powerful. ICSM preserves holistic report-level coherence that disease-centric alignment alone cannot guarantee. Configurations relying only on node-level alignment (row 9 in Table~\ref{tab:ablation}) improve finding-level precision but do not match the full model, confirming that global semantic structure provides a necessary foundation for disease-conditioned alignment.

\subsubsection{Impact of Disease-Centric Alignment}

Introducing \textbf{ICDM} alone yields a modest but consistent improvement, confirming that shifting cross-modal discrimination from the report level to disease-specific nodes is beneficial even with hard one-hot targets. Adding \textbf{DCNR} further improves performance by relaxing strict one-to-one targets: it allows studies sharing the same finding to contribute soft positive signal, reducing false negatives that are inherent in medical contrastive learning. The largest gain comes from \textbf{IMSC}: BLEU-4 increases from 0.134 to 0.139 and ROUGE from 0.294 to 0.299. This reveals that disease-specific supervision units require not only cross-modal discrimination but also intra-modal separability. These objectives are complementary: ICDM provides the discriminative backbone, DCNR adapts it to clinical reality, and IMSC regularizes the intra-modal feature space. More broadly, comparing rows 3 vs.\ 5 and 4 vs.\ 7, adding disease-conditioned soft regularization (DCNR/DCSR) consistently improves over hard-matching counterparts at both levels, confirming that it refines discriminative alignment rather than replacing it.

\subsubsection{Impact of Knowledge Graph}

The value of the knowledge graph lies in how it structures supervision units, not merely in the number of nodes it introduces. We verify this from two perspectives: ontology extensibility and graph structure. As shown in Table~\ref{tab:graph_ablation}, extending the 14-disease CheXpert graph to a 25-disease graph constructed from RadGraph-derived~\cite{jain2021radgraph} entity statistics yields consistent improvements on both lexical and clinical metrics. This indicates that the framework can naturally absorb a broader disease ontology; the gains are modest, as expected given the long-tailed frequency distribution of the newly added entities. We further compare our anatomically informed graph with a random graph and an all-connect graph. Both variants underperform the original design. Random connectivity removes clinically meaningful priors, while full connectivity weakens discriminability by introducing non-informative relations between unrelated findings. The graph therefore helps by providing a clinically structured training-time prior over supervision units, not by adding parameters.

\begin{table}[h]
\centering
\caption{Ablation on graph design. ``25-Disease'' extends the ontology with RadGraph-derived entities. ``Random'' uses a random adjacency matrix; ``All-connect'' links every node.}
\label{tab:graph_ablation}
\resizebox{\linewidth}{!}{
\begin{tabular}{lccccc}
\hline
\textbf{Method} & \textbf{B@4} & \textbf{ROUGE-L} & \textbf{RadGraph F1} $\uparrow$ & \textbf{BERTScore} $\uparrow$ & \textbf{RadCliQ} $\downarrow$ \\
\hline
Random Graph & 0.137 & 0.298 & 0.198 & 0.417 & 1.167 \\
All-connect Graph & 0.139 & 0.298 & 0.201 & 0.419 & 1.158 \\
Ours & 0.143 & \textbf{0.305} & 0.206 & 0.422 & 1.145 \\
25-Disease Extended Graph & \textbf{0.144} & \textbf{0.305} & \textbf{0.209} & \textbf{0.425} & \textbf{1.138} \\
\hline
\end{tabular}
}
\end{table}

\subsubsection{Impact of Hyperparameters}
We test the impact of hyperparameters $\lambda_1$ and $\lambda_2$ in Table~\ref{tab:ablation_hype}. When the weight $\lambda_2$ of Disease-Centric Alignment increases to 2, our model improves further, reinforcing the importance of disease-specific supervision units. 

\begin{table}[h]
\caption{Ablation study for hyperparameters: $\lambda_1$ and $\lambda_2$.}
\centering
{\small
\begin{tabular}{@{}c|c|c|c|c|c|c|c@{}}
\hline
\textbf{$\lambda_1$}    & \textbf{$\lambda_2$}       & \textbf{B@1}              & \textbf{B@2}              & \textbf{B@3}              & \textbf{B@4}                 & \textbf{METEOR}              & \textbf{ROUGE}       \\ \hline
1                       & 1                          & 0.426            & 0.277            & 0.192            & 0.139               & 0.167               & 0.298       \\
0.5                     & 1                          & 0.426            & 0.276            & 0.191            & 0.138               & 0.167               & 0.298       \\
1.5                     & 1                          & 0.426            & 0.276            & 0.190            & 0.137               & 0.166               & 0.296       \\
2                       & 1                          & 0.425            & 0.276            & 0.190            & 0.137               & 0.166               & 0.296       \\
1                       & 0.5                        & 0.424            & 0.277            & 0.192            & 0.139               & 0.165               & 0.299       \\
1                       & 1.5                        & 0.425            & 0.276            & 0.191            & 0.138               & 0.166               & 0.299       \\
1                       & 2                          & \textbf{0.427}            & \textbf{0.279}            & \textbf{0.194}            & \textbf{0.143}               & \textbf{0.167}               & \textbf{0.305}       \\
\hline
\end{tabular}
}
\label{tab:ablation_hype}
\end{table}
%
\begin{figure}[!h]
    \centering
    \includegraphics[width=\linewidth]{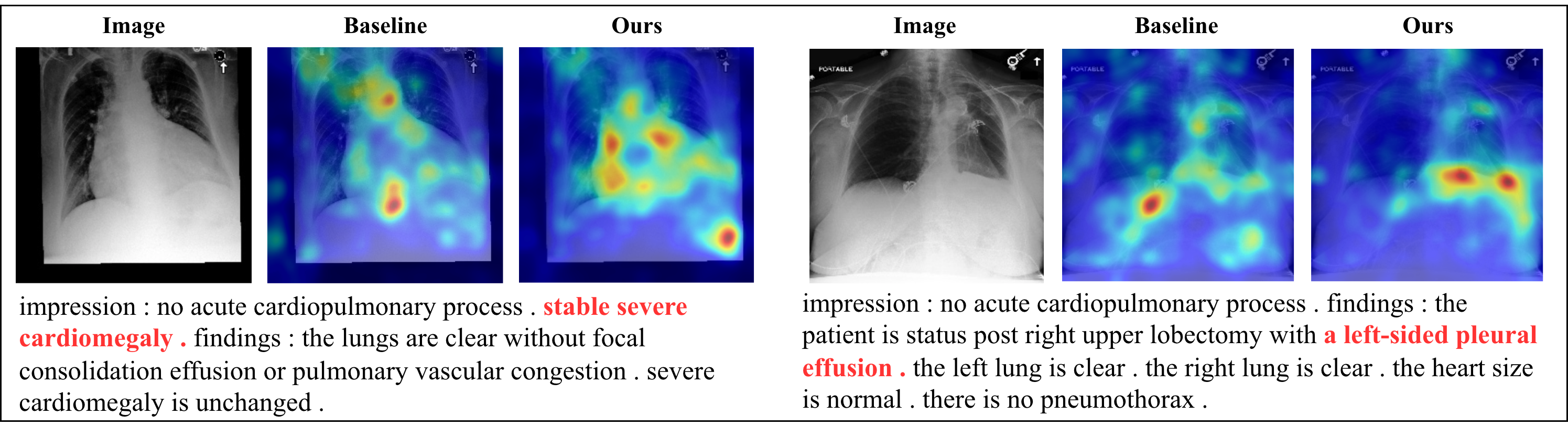}
    \caption{Case-level vision interpretability via heatmaps.}
    \label{fig:cam}
\end{figure}

\subsection{Qualitative Analysis}

Figure~\ref{fig:Qulitative} shows representative reports from the baseline, a variant with only \textbf{Global Clinical Semantic Alignment}, and our full model. The baseline captures normal features but misses abnormalities. Adding global alignment recovers coarse findings such as ``edema,'' but finer details remain incomplete. The full model further generates disease-specific descriptions such as ``cardiac silhouette is mildly enlarged,'' indicating that \textbf{Disease-Centric Alignment} improves finding-level precision. Figure~\ref{fig:cam} visualizes case-level attention: our model focuses on clinically relevant regions (e.g., cardiac silhouette for cardiomegaly, costophrenic angle for pleural effusion), confirming that the framework improves disease-conditioned alignment rather than relying on language priors.

\section{Conclusions}
\label{sec:conclusions}

This work argues that the current challenge of radiology report generation lies not in language generation capacity, but in the granularity at which supervision is imposed. We reformulate image-report supervision in RRG as a hierarchical clinical alignment problem and propose \textbf{Graph-Supervised Hierarchical Clinical Alignment}, which integrates \textbf{Disease-Centric Alignment} for fine-grained disease-level correspondence with \textbf{Global Clinical Semantic Alignment} for report-level coherence. The resulting framework improves both generation quality and clinical faithfulness across three benchmarks, while using the knowledge graph only as a training-time structural prior that specifies how supervision should be factorized, not how inference should be performed.



\bibliographystyle{ACM-Reference-Format}
\bibliography{main}


\end{document}